\pdfoutput=1
\documentclass[10pt]{article}

\usepackage[utf8]{inputenc}
\usepackage[T1]{fontenc}
\usepackage[a4paper,top=2cm,bottom=2cm,left=1.8cm,right=1.8cm,columnsep=0.7cm]{geometry}
\usepackage{times}
\usepackage{amsmath,amssymb}
\usepackage{graphicx}
\usepackage{booktabs}
\usepackage{array}
\usepackage{multirow}
\usepackage{url}
\usepackage[font=small,labelfont=bf]{caption}
\usepackage{enumitem}
\usepackage{microtype}
\usepackage[hidelinks]{hyperref}

\setlist[itemize]{leftmargin=*,topsep=0.3em,itemsep=0.2em}

\newcolumntype{L}[1]{>{\raggedright\arraybackslash}p{#1}}

\makeatletter
\renewcommand{\@maketitle}{%
  \begin{center}
    {\LARGE\bfseries \@title \par}
    \vskip 0.9em
    {\large
      \lineskip 0.5em
      \begin{tabular}[t]{c}
        \@author
      \end{tabular}\par}
  \end{center}
  \par\vskip 0.8em}
\makeatother

\newcommand{\paperfigsingle}[1]{\includegraphics[width=\linewidth,height=0.22\textheight,keepaspectratio]{#1}}
\newcommand{\paperfigwide}[1]{\includegraphics[width=0.98\textwidth,height=0.32\textheight,keepaspectratio]{#1}}
\newcommand{\paperfiglarge}[1]{\includegraphics[width=\linewidth,height=0.26\textheight,keepaspectratio]{#1}}

\begin{document}

\title{A Risk-Field Enhanced Closed-Loop Digital Twin Framework for Autonomous Driving Safety Validation}

\author{Yongzhi Liu\\
School of Mechanical Engineering\\
Southeast University\\
Nanjing, China\\
\texttt{yongzhi.liu@seu.edu.cn}}

\date{}

\twocolumn[
\begin{@twocolumnfalse}
\maketitle

\begin{abstract}
Autonomous driving systems require reliable safety validation before real-world deployment. However, large-scale road testing is costly, difficult to reproduce, and inefficient for exposing rare safety-critical scenarios. Conventional simulation improves repeatability, but an offline simulator alone cannot continuously connect physical traffic states, virtual reconstruction, algorithm evaluation, and scenario evolution. This paper proposes a risk-field enhanced closed-loop digital twin framework for autonomous driving safety validation. The framework integrates physical data acquisition, data synchronization, virtual twin reconstruction, risk-aware scenario generation, autonomous driving algorithm evaluation, and safety analysis. A driving risk field is introduced as a unified intermediate representation to describe obstacle, lane-departure, road-boundary, time-to-collision, and comfort-related risks around the ego vehicle. The risk field ranks high-risk scenarios in the digital twin scenario library and provides dense safety guidance for reinforcement learning-based driving policies. A simulation-style evaluation protocol is designed to compare conventional reinforcement learning baselines, risk-penalty baselines, and the proposed risk-field guided method. The study indicates that embedding explicit risk structure into digital twins can make autonomous driving validation more targeted, interpretable, and reusable, while its practical effectiveness remains bounded by model fidelity, risk calibration, and sim-to-real transfer.
\end{abstract}

\noindent\textbf{Keywords:} Digital twin, autonomous driving, risk field, scenario generation, reinforcement learning, safety validation.

\vspace{1.1em}
\end{@twocolumnfalse}
]

\section{Introduction}
\label{sec:introduction}

Autonomous driving is a representative application of artificial intelligence, robotics, vehicle engineering, and cyber-physical systems. A deployed autonomous vehicle must perceive dynamic objects, predict interactive behaviours, plan feasible trajectories, and execute control commands under real-time constraints. These functions must remain reliable under complex road geometry, weather changes, traffic density variations, sensor noise, occlusions, and unpredictable human driving behaviours.

Safety validation is therefore a central bottleneck in autonomous driving development. Real-road testing provides indispensable evidence, but rare-event safety cannot be demonstrated efficiently by mileage accumulation alone \cite{kalra2016driving}. Many hazardous cases, such as sudden cut-in, emergency braking, occluded pedestrian crossing, sensor failure, and low-friction road conditions, are difficult to collect at scale and unsafe to reproduce physically. Simulation platforms such as CARLA can generate controllable traffic scenes and support repeatable evaluation \cite{dosovitskiy2017carla}. However, the value of simulation depends on whether virtual cases remain predictive of real driving behaviour \cite{hu2024simulation}.

Digital twin technology provides a promising way to connect real-world data and virtual validation. A digital twin is not merely a static model, but a virtual representation that maintains a data relationship with physical entities throughout operation or lifecycle \cite{grieves2017,kritzinger2018,fuller2020}. In autonomous driving, a digital twin can represent vehicles, road infrastructure, sensors, traffic participants, high-definition maps, communication states, and scenario libraries. It can reconstruct real operational cases, evaluate autonomous driving algorithms, and feed failure information back to scenario generation and model updating \cite{yu2022avdt,wang2024implementation,liang2025dttfsim}.

However, a practical autonomous driving digital twin should do more than reproduce scenes. It should identify which parts of a scene are safety-critical and convert this information into actionable validation pressure. Existing simulation and scenario-based testing methods often rely on manually specified scenario parameters or sparse failure indicators \cite{riedmaier2020survey,neurohr2021criticality}. These indicators are useful for post-test evaluation, but they provide limited guidance before a failure occurs. A digital twin that explicitly models the spatial and temporal distribution of driving risk can expose unsafe regions earlier and support more interpretable closed-loop validation.

To address this issue, this paper proposes a risk-field enhanced closed-loop digital twin framework for autonomous driving safety validation. The main idea is to introduce a driving risk field into the digital twin loop. The risk field describes potential danger around the ego vehicle using obstacle proximity, relative motion, lane deviation, road-boundary distance, time-to-collision (TTC), and comfort-related constraints. It connects three core functions: scenario generation, risk evaluation, and reinforcement learning-based policy training.

The main contributions of this paper are summarized as follows:
\begin{itemize}
    \item A closed-loop digital twin framework is developed for autonomous driving safety validation by connecting physical data, virtual reconstruction, algorithm evaluation, risk assessment, and scenario evolution.
    \item A driving risk field is introduced as a unified representation that links high-risk scenario generation, safety evaluation, and risk-guided policy learning.
    \item A risk-field enhanced reinforcement learning formulation is designed to provide dense safety guidance for autonomous driving policies.
    \item A simulation-style evaluation protocol is provided to analyze success rate, collision rate, risk score, route completion, training stability, and control smoothness.
\end{itemize}

\begin{figure*}[!t]
    \centering
    \paperfigwide{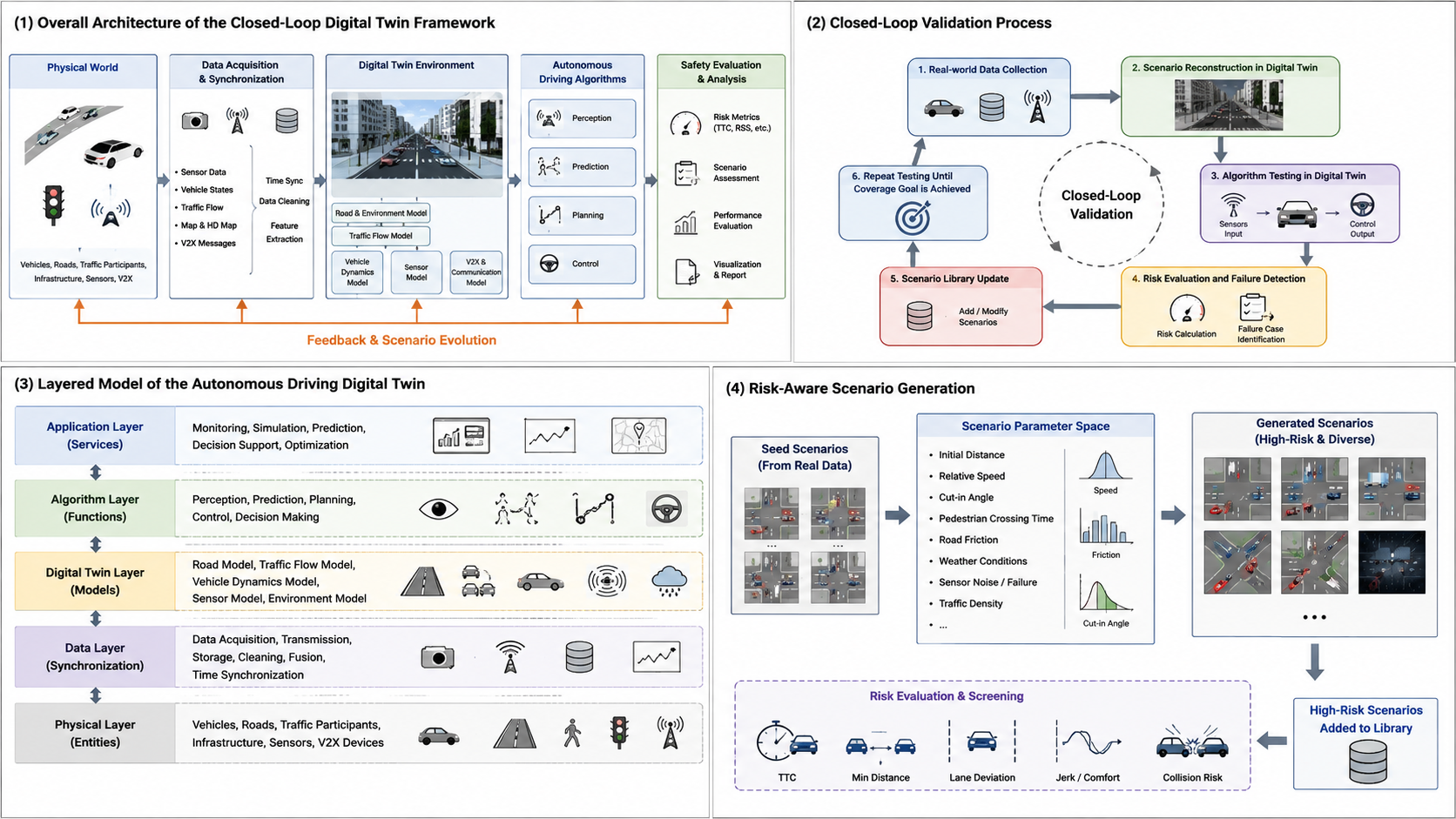}
    \caption{Overall architecture of the proposed closed-loop digital twin framework. Physical-world information is synchronized into a virtual twin, where autonomous driving algorithms are evaluated, safety metrics are calculated, and high-risk scenarios are updated for iterative validation.}
    \label{fig:overall_framework}
\end{figure*}

\section{Related Work}
\label{sec:related_work}

\subsection{Digital Twin Technology}

The digital twin concept was introduced to describe a virtual representation that is linked to a physical object or system throughout its lifecycle \cite{grieves2017}. Later studies clarified the distinction among digital models, digital shadows, and digital twins according to the direction and automation of data flows \cite{kritzinger2018}. Digital twins have been studied in smart manufacturing, industrial systems, smart cities, and intelligent transportation systems \cite{tao2017shopfloor,fuller2020,jafari2023review}. Compared with traditional simulation, digital twin technology emphasizes continuous data synchronization, physical-virtual consistency, lifecycle management, prediction, and decision support.

\subsection{Digital Twins for Autonomous Driving}

In autonomous driving, digital twins can represent vehicles, roads, traffic participants, sensors, infrastructure, and communication systems. Yu et al. described autonomous vehicle digital twins as a practical paradigm for autonomous driving system development \cite{yu2022avdt}. Hu et al. connected simulation, Sim2Real transfer, digital twins, and parallel intelligence for autonomous driving validation \cite{hu2024simulation}. Wang et al. demonstrated how roadside units, edge computing, and cloud processing can support autonomous driving digital twins \cite{wang2024implementation}. DTTF-Sim showed that digital twins can support continuous autonomous driving testing by reconstructing realistic traffic flow \cite{liang2025dttfsim}. These studies motivate digital twins as validation infrastructure, but risk representation inside the validation loop remains insufficiently explored.

\subsection{Scenario-Based Safety Validation}

Scenario-based testing evaluates automated vehicles on representative and safety-critical traffic situations rather than only on aggregate mileage. Riedmaier et al. surveyed scenario-based safety assessment methods for automated vehicles \cite{riedmaier2020survey}. Neurohr et al. discussed criticality analysis for verification and validation of automated vehicles \cite{neurohr2021criticality}. CommonRoad provides composable benchmarks for motion planning on roads \cite{althoff2017commonroad}. Data-driven simulation methods such as AADS show that real-world driving data can improve the realism of virtual testing \cite{li2019aads}. Since real-world critical events are rare and expensive to collect, safety-critical scenario generation has become an important research topic \cite{hanselmann2022king}.

\subsection{Reinforcement Learning for Autonomous Driving}

Reinforcement learning provides a general framework for sequential decision-making and autonomous control \cite{sutton2018}. Algorithms such as proximal policy optimization (PPO) \cite{schulman2017ppo} and soft actor-critic (SAC) \cite{haarnoja2018sac} are widely used in continuous control. In autonomous driving, reinforcement learning has been studied for lane keeping, intersection navigation, speed control, and decision-making in simulation environments \cite{guti2022carla}. However, reinforcement learning often suffers from unsafe exploration and sparse rewards. Dense reinforcement learning and safety-critical scenario search show that learning-based methods can actively expose failure cases for autonomous vehicle validation \cite{feng2023dense}. Recent world-model-based and risk-aware dual-policy methods also demonstrate promising performance for safe and adaptive highway autonomous driving \cite{liu2025riskaware}. This motivates the use of risk fields to provide dense and interpretable safety guidance.

\section{Proposed Framework and Methodology}
\label{sec:method}

\subsection{Overall Architecture}

Fig.~\ref{fig:overall_framework} shows the overall architecture of the proposed risk-field enhanced closed-loop digital twin framework. The framework contains five layers: physical layer, data layer, virtual twin layer, autonomous driving algorithm layer, and safety evaluation layer. The physical layer includes autonomous vehicles, road infrastructure, traffic participants, and sensors. The data layer performs acquisition, cleaning, synchronization, storage, and V2X or edge-cloud communication. The virtual twin layer reconstructs vehicle dynamics, traffic flow, sensor observations, and virtual scenario environments. The algorithm layer evaluates perception, prediction, planning, and control modules. The safety evaluation layer calculates risk metrics and updates the scenario library.

\begin{figure*}[!t]
    \centering
    \paperfigwide{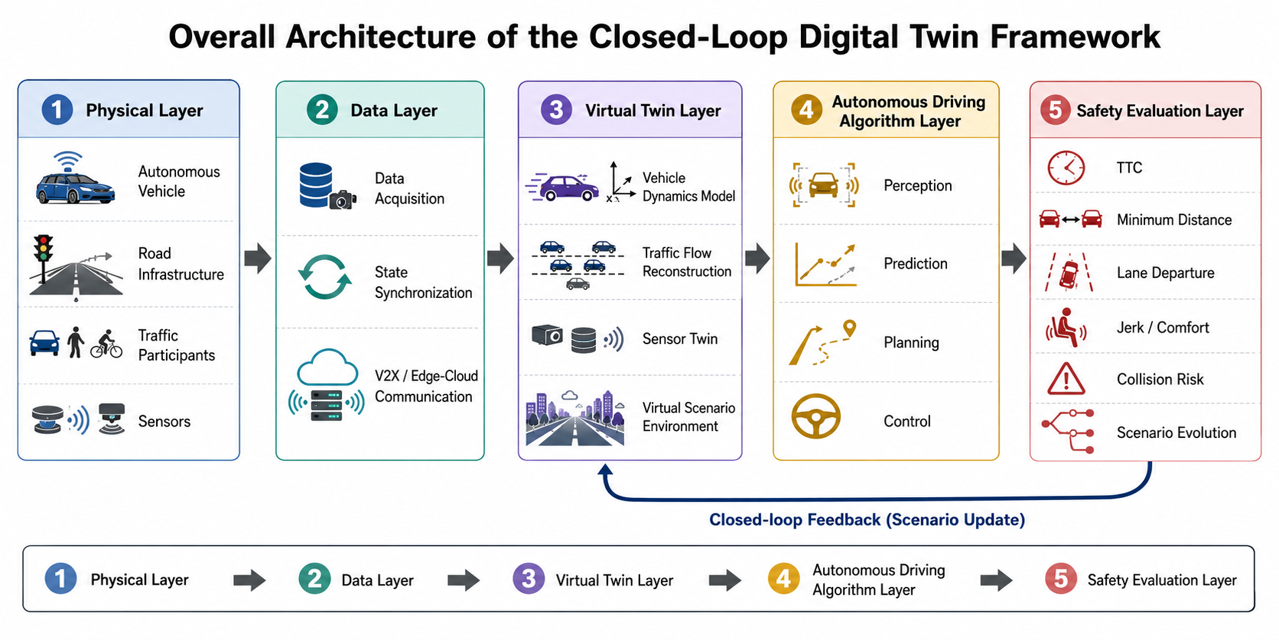}
    \caption{Closed-loop digital twin framework for autonomous driving validation. The framework links the physical layer, data layer, virtual twin layer, autonomous driving algorithm layer, and safety evaluation layer through bidirectional feedback.}
    \label{fig:cldt_framework}
\end{figure*}

\begin{figure}[!t]
    \centering
    \paperfigsingle{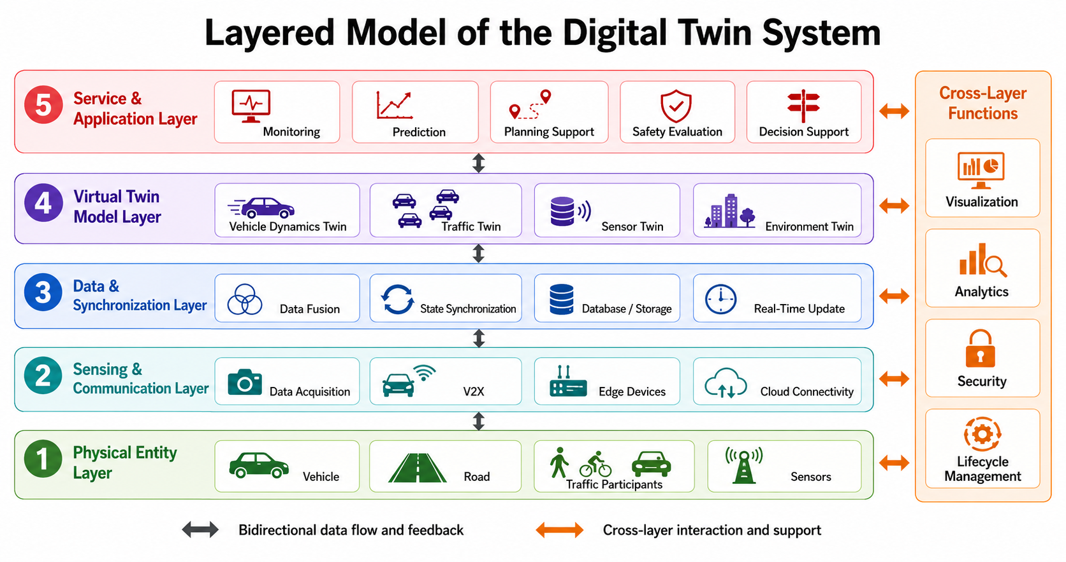}
    \caption{Layered model of the autonomous driving digital twin system. Physical entities, communication, data synchronization, virtual models, and service applications are connected by bidirectional data flow and cross-layer support.}
    \label{fig:layered_model}
\end{figure}

The closed-loop digital twin framework is detailed in Fig.~\ref{fig:cldt_framework}, which illustrates how the five layers are connected through bidirectional data and feedback flows. The layered model in Fig.~\ref{fig:layered_model} further clarifies the function of each layer. The physical entity layer anchors the twin to vehicles, roads, traffic participants, infrastructure, and sensors. The sensing and communication layer collects information through onboard sensors, roadside units, V2X devices, edge nodes, and cloud services. The data synchronization layer performs fusion, time alignment, storage, and real-time update. The virtual twin model layer supports vehicle dynamics twins, traffic twins, sensor twins, and environment twins. The service layer provides monitoring, prediction, planning support, safety evaluation, and decision support.

\subsection{Closed-Loop Validation Process}

\begin{figure}[!t]
    \centering
    \paperfigsingle{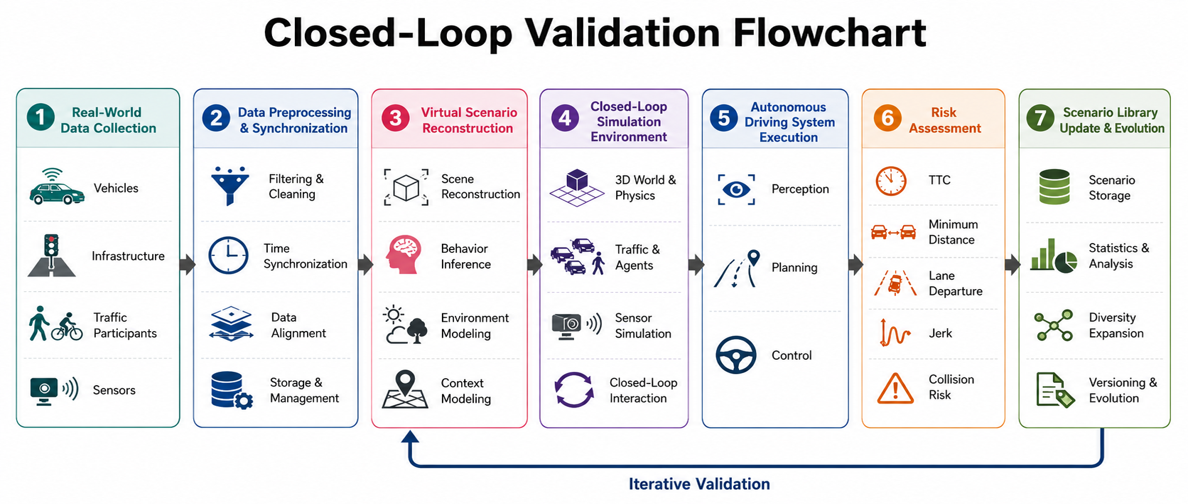}
    \caption{Closed-loop validation process for autonomous driving digital twins. Real-world data are transformed into virtual scenarios, algorithms are tested in simulation, and risk assessment updates the scenario library.}
    \label{fig:validation_flow}
\end{figure}

The closed-loop validation flow is shown in Fig.~\ref{fig:validation_flow}. It starts with real-world data collection from vehicles, infrastructure, traffic participants, and sensors. These data are filtered, cleaned, synchronized, and aligned to a consistent time base. The digital twin then reconstructs virtual scenarios with road topology, traffic agents, vehicle dynamics, and sensor models. The autonomous driving algorithm is evaluated in the reconstructed environment. Risk metrics are calculated to identify failure cases, and the scenario library is updated for iterative validation.

This process differs from ordinary offline simulation in two aspects. First, virtual scenarios are not isolated handcrafted tests, but are derived from physical data, scenario libraries, and failure feedback. Second, safety evaluation is not only a final score; it shapes scenario evolution and can guide policy learning through risk-field feedback. In this way, a collision, near miss, excessive lane departure, or uncomfortable control action can be converted into a reusable validation case.

The data layer is critical for making this loop credible. Vehicle controller area network signals, perception outputs, traffic-light states, roadside observations, and map information usually arrive at different frequencies. The digital twin therefore stores timestamped records and reconstructs a synchronized state before each virtual replay. This avoids evaluating a driving policy on a scene in which the ego vehicle, surrounding agents, and signal phases are physically inconsistent.

\subsection{Risk-Aware Scenario Generation}

\begin{figure}[!t]
    \centering
    \paperfigsingle{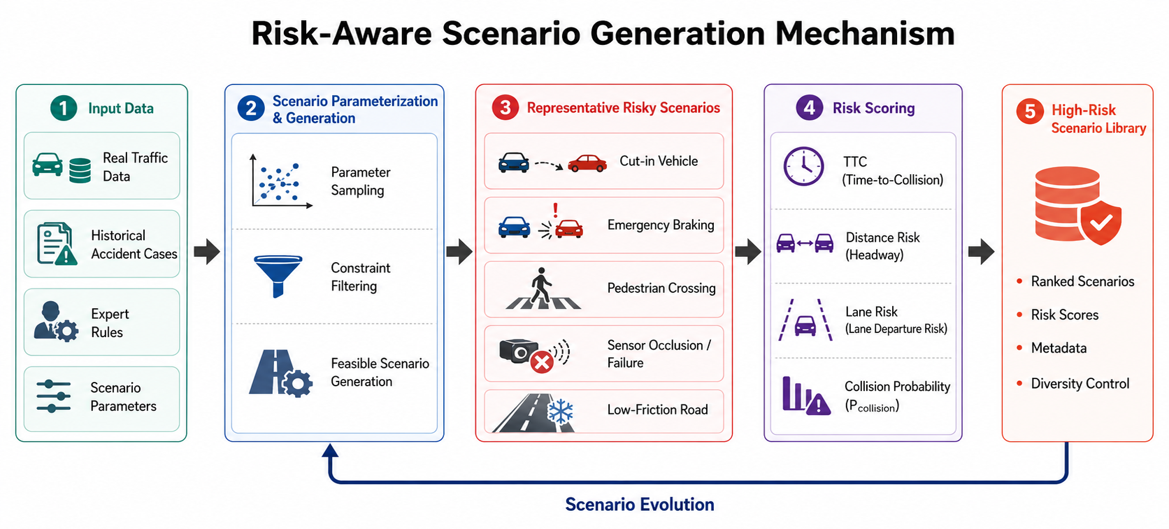}
    \caption{Risk-aware scenario generation mechanism. Real traffic data, accident cases, expert rules, and scenario parameters are transformed into representative risky scenarios and ranked by risk scores before entering the scenario library.}
    \label{fig:scenario_generation}
\end{figure}

Fig.~\ref{fig:scenario_generation} shows the risk-aware scenario generation mechanism. The input sources include real traffic data, historical accident cases, expert rules, and scenario parameters. The generation module samples parameters, filters infeasible combinations, and constructs representative high-risk scenarios, such as cut-in vehicles, emergency braking, pedestrian crossing, sensor occlusion, and low-friction roads. Each generated scenario is scored by risk metrics and stored in the high-risk scenario library.

The key difference from random scenario sampling is that risk evaluation is embedded in the generation loop. A scenario is not selected only because it is visually plausible; it is selected because it stresses a measurable safety dimension, such as TTC, minimum distance, lane departure, collision probability, or rule violation potential. This design makes scenario generation a validation tool rather than a purely visual scene construction process.

\subsection{Driving Risk Field}

To quantify the safety state around the ego vehicle, this paper defines a driving risk field. Let $p=(x,y)$ denote a spatial position in the local driving coordinate frame at time $t$. The total risk field is defined as
\begin{equation}
\begin{aligned}
\mathcal{R}(p,t)={}&
\alpha_o\mathcal{R}_o(p,t)+
\alpha_l\mathcal{R}_l(p,t) \\
&+\alpha_b\mathcal{R}_b(p,t)+
\alpha_c\mathcal{R}_c(p,t),
\end{aligned}
\end{equation}
where $\mathcal{R}_o$, $\mathcal{R}_l$, $\mathcal{R}_b$, and $\mathcal{R}_c$ represent obstacle risk, lane risk, road-boundary risk, and comfort-related risk, respectively. The coefficients $\alpha_o$, $\alpha_l$, $\alpha_b$, and $\alpha_c$ control their relative importance.

For a surrounding vehicle $i$, obstacle risk can be modeled as
\begin{equation}
\mathcal{R}_{o,i}(p,t)=
\exp\left(-\frac{d_i^2(p,t)}{2\sigma_d^2}\right)
\left(1+\lambda_v \Delta v_i(t)\right),
\end{equation}
where $d_i(p,t)$ is the distance between position $p$ and vehicle $i$, $\sigma_d$ controls the spatial influence range, $\Delta v_i(t)$ denotes the relative velocity term, and $\lambda_v$ is the velocity-risk coefficient.

The lane-departure risk is calculated by the lateral deviation from the lane center:
\begin{equation}
\mathcal{R}_l(t)=
\left(\frac{|e_y(t)|}{w_l/2}\right)^2,
\end{equation}
where $e_y(t)$ is the lateral deviation and $w_l$ is the lane width. A larger deviation indicates a higher probability of lane departure or unsafe lateral behaviour.

\begin{figure}[!t]
    \centering
    \paperfiglarge{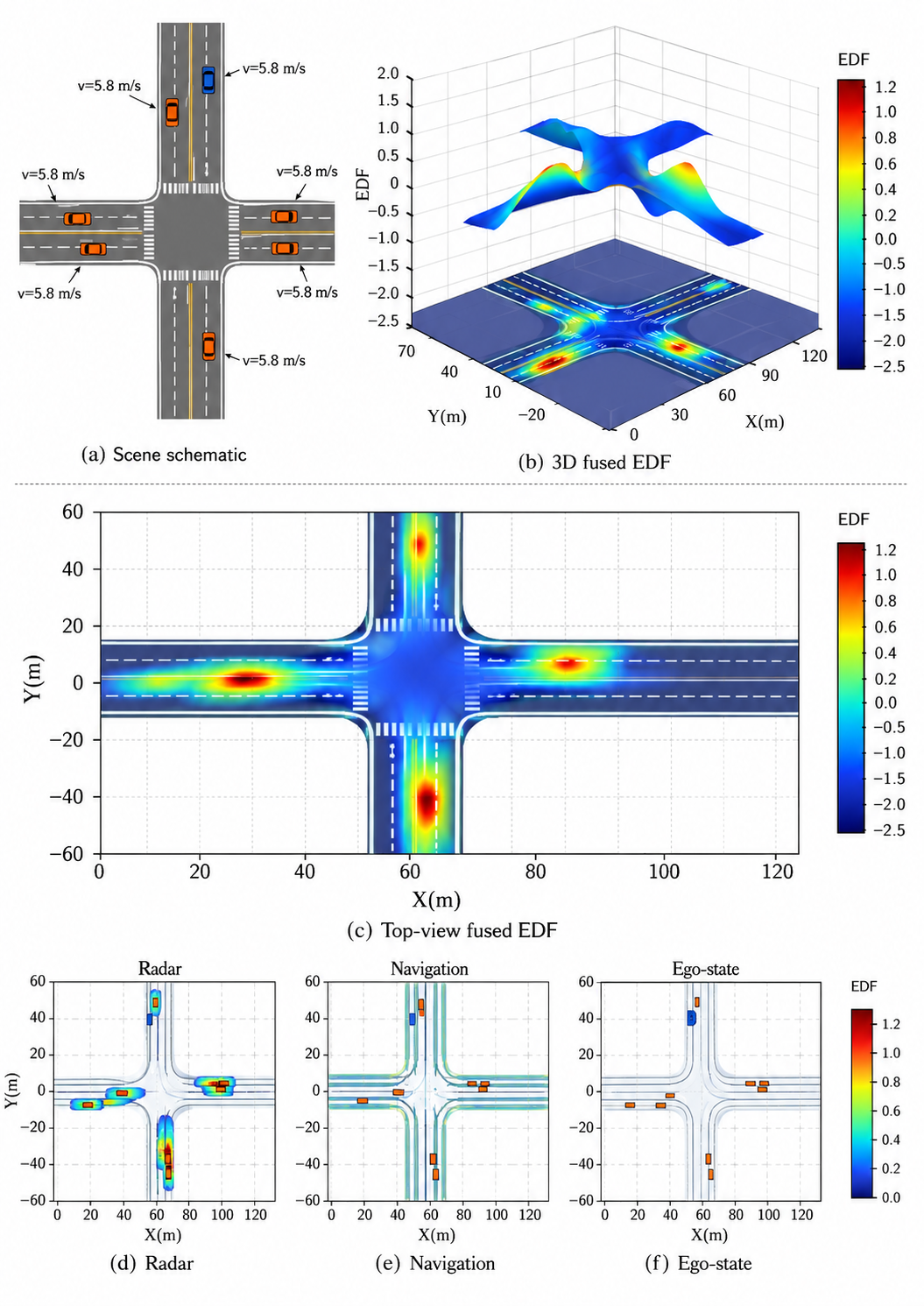}
    \caption{Driving risk field at a multi-vehicle intersection. The field represents spatially varying risk around lanes, conflict points, and nearby vehicles, which enables earlier detection of unsafe regions than a binary collision indicator.}
    \label{fig:risk_field}
\end{figure}

For scenario-library update, the field is reduced to a scalar risk score:
\begin{equation}
S_{\mathrm{risk}}=
\beta_1 S_{\mathrm{TTC}}+
\beta_2 S_{\mathrm{dist}}+
\beta_3 S_{\mathrm{lane}}+
\beta_4 S_{\mathrm{jerk}}+
\beta_5 S_{\mathrm{col}},
\end{equation}
where $S_{\mathrm{TTC}}$ is the time-to-collision risk, $S_{\mathrm{dist}}$ is the minimum-distance risk, $S_{\mathrm{lane}}$ is the lane-departure risk, $S_{\mathrm{jerk}}$ is the comfort risk, and $S_{\mathrm{col}}$ is the collision risk. Scenarios with larger $S_{\mathrm{risk}}$ are prioritized for replay, mutation, and regression testing.

The coefficients in the risk score should be treated as validation parameters rather than arbitrary tuning constants. In a practical system, their values should be calibrated from traffic rules, vehicle limits, and accident statistics. For example, TTC and minimum-distance terms should dominate in dense interactions, while lane and comfort terms should become more important when the vehicle is far from direct conflict. This calibrated design prevents the risk field from becoming either too conservative or too insensitive to early danger.

\subsection{Risk-Guided Reinforcement Learning}

The autonomous driving task in the digital twin is formulated as a Markov decision process (MDP), defined by $(\mathcal{S}, \mathcal{A}, P, r, \gamma)$. The state $s_t$ contains ego-vehicle state, road state, surrounding traffic state, and risk-field features:
\begin{equation}
\begin{split}
s_t=[&v_t,a_t,\psi_t,e_y,e_\psi,d_f,\Delta v_f,\\
&\rho_{\mathrm{lane}},\rho_{\mathrm{obs}},\rho_{\mathrm{TTC}}],
\end{split}
\end{equation}
where $v_t$ is ego speed, $a_t$ is ego acceleration, $\psi_t$ is heading angle, $e_y$ is lateral deviation, $e_\psi$ is heading error, $d_f$ is front distance, $\Delta v_f$ is relative speed to the front vehicle, and $\rho_{\mathrm{lane}}$, $\rho_{\mathrm{obs}}$, and $\rho_{\mathrm{TTC}}$ denote lane, obstacle, and TTC risk features.

The action $u_t$ is defined as continuous vehicle control:
\begin{equation}
u_t=[\delta_t, throttle_t, brake_t],
\end{equation}
where $\delta_t$ is steering angle, $throttle_t$ is throttle command, and $brake_t$ is braking command.

The reward combines task achievement, speed tracking, risk avoidance, lane keeping, comfort, and terminal safety penalties:
\begin{equation}
\begin{split}
r_t ={}&
w_p r_{\mathrm{progress}}
+w_v r_{\mathrm{velocity}}
-w_r \bar{\mathcal{R}}_t
-w_l |e_y(t)| \\
&-w_j |j_t|
-w_c \mathbb{I}_{\mathrm{collision}}
-w_o \mathbb{I}_{\mathrm{offroad}},
\end{split}
\end{equation}
where $r_{\mathrm{progress}}$ encourages forward progress, $r_{\mathrm{velocity}}$ encourages reasonable speed tracking, $\bar{\mathcal{R}}_t$ is the risk sampled around the ego vehicle or along a candidate trajectory, $j_t$ is jerk, $\mathbb{I}_{\mathrm{collision}}$ is the collision indicator, and $\mathbb{I}_{\mathrm{offroad}}$ is the off-road indicator.

Compared with sparse collision penalties, the risk-field term provides dense safety guidance before collision occurs. It discourages the agent from entering high-risk regions and helps the policy learn smoother avoidance behaviour during training. This makes the reward function more suitable for safety validation than a reward defined only by task success and terminal collision.

\section{Experimental Setup}
\label{sec:experimental_setup}

\subsection{Scenario Configuration}

The evaluation is organized as a simulation-style validation protocol. The virtual environment contains urban roads, intersections, surrounding vehicles, pedestrians, and different traffic densities. Five representative high-risk scenario types are considered: cut-in vehicle, emergency braking, pedestrian crossing, sensor occlusion, and low-friction road. These scenario types cover both interaction risks and perception-related risks.

The scenario generation module produces parameterized variants for each type. For cut-in cases, the key parameters include lateral gap, longitudinal gap, relative velocity, and cut-in timing. For emergency braking, the parameters include initial headway, braking intensity, and road friction. For pedestrian crossing, the parameters include crossing position, speed, occlusion level, and ego speed. Sensor occlusion and low-friction scenarios are used to test robustness under degraded perception and control conditions.

\subsection{Baselines and Variants}

The baseline methods include PPO, SAC, DreamerV3, R2D2-style recurrent reinforcement learning, and a large language model-assisted driving agent, following the comparison groups shown in the training curves. The proposed method is compared with risk-agnostic reinforcement learning and variants that remove risk-aware scenario generation or risk-field reward weighting.

Three variants are used for ablation. The base variant uses a conventional task reward without explicit risk-field guidance. The risk-only variant adds risk penalties but does not update the scenario library through risk-aware generation. The full variant combines closed-loop digital twin validation, risk-aware scenario generation, and risk-field reward shaping. This design tests whether the improvement comes from a single reward term or from the coupled digital twin loop.

\subsection{Evaluation Metrics}

The evaluation focuses on seven metrics: reward, success rate, route completion, collision rate, average speed, steering smoothness, and longitudinal smoothness. Reward measures the overall learning signal. Success rate and route completion measure task achievement. Collision rate and risk score measure safety. Average speed prevents a trivial solution in which the vehicle avoids danger by stopping. Steering and longitudinal smoothness measure whether the learned policy is suitable for passenger-oriented deployment.

\begin{table}[!t]
\centering
\caption{Evaluation metrics used for closed-loop safety validation}
\label{tab:metrics}
\footnotesize
\begin{tabular}{@{}L{0.30\linewidth} L{0.58\linewidth}@{}}
\toprule
Metric & Validation meaning \\
\midrule
Reward & Overall learning signal combining progress, safety, and comfort. \\
Success rate & Fraction of episodes that complete the driving task without terminal failure. \\
Route completion & Percentage of the reference route completed by the ego vehicle. \\
Collision rate & Frequency of contacts with vehicles, pedestrians, or obstacles. \\
Average speed & Whether the policy remains useful rather than overly conservative. \\
Steering smoothness & Lateral control stability and passenger comfort. \\
Longitudinal smoothness & Acceleration and braking stability under interaction pressure. \\
\bottomrule
\end{tabular}
\end{table}

These metrics are intentionally complementary. A method that maximizes success rate but produces unstable steering is not acceptable for autonomous driving. Similarly, a method that avoids collision by stopping is not useful for real deployment. Therefore, the evaluation emphasizes balanced performance rather than a single headline score.

\section{Results and Analysis}
\label{sec:results}

\subsection{Training Behaviour}

\begin{figure}[!t]
    \centering
    \paperfiglarge{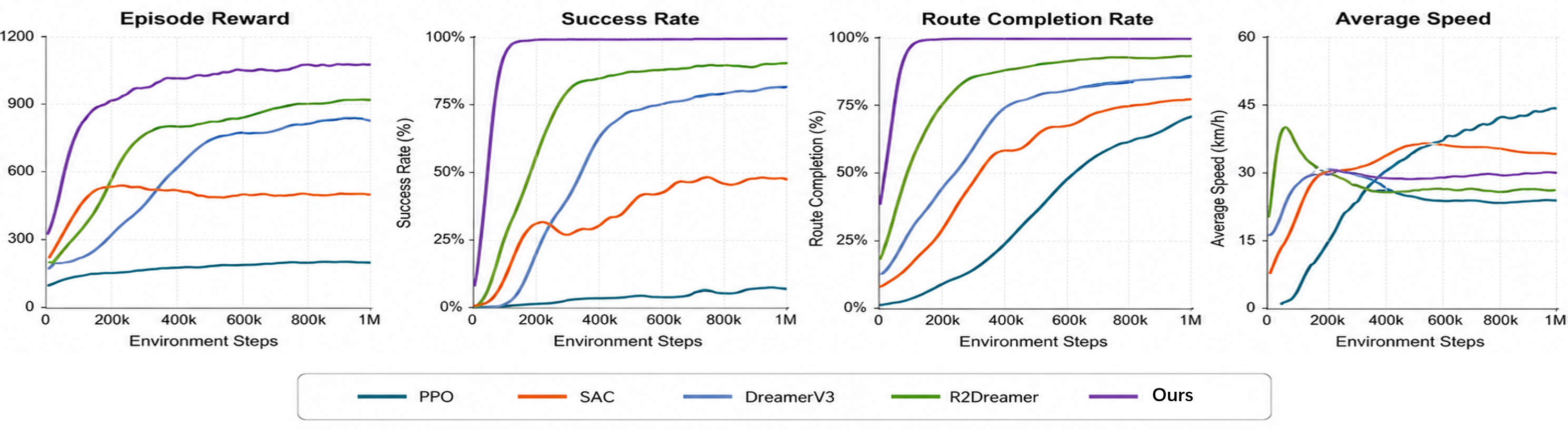}
    \caption{Training curves for reinforcement learning baselines and the proposed risk-field enhanced method. The curves compare reward, success rate, route completion, and speed over environment steps.}
    \label{fig:training_curves}
\end{figure}

Fig.~\ref{fig:training_curves} reports the training behaviour across episode reward, success rate, route completion, and average speed. The proposed risk-field enhanced method converges to higher reward and maintains a more stable success trend than the baselines. The route completion curve also improves more consistently, indicating that the dense risk signal does not merely stop the vehicle from taking risks, but helps it complete the driving task more reliably.

The main interpretation is that risk guidance changes the learning problem from delayed failure avoidance to continuous risk shaping. A standard reward must wait until collision, off-road behaviour, or task failure occurs before applying a strong penalty. The proposed field provides a smoother safety gradient, which makes unsafe regions visible earlier in the episode.

\subsection{Ablation Analysis}

\begin{figure}[!t]
    \centering
    \paperfiglarge{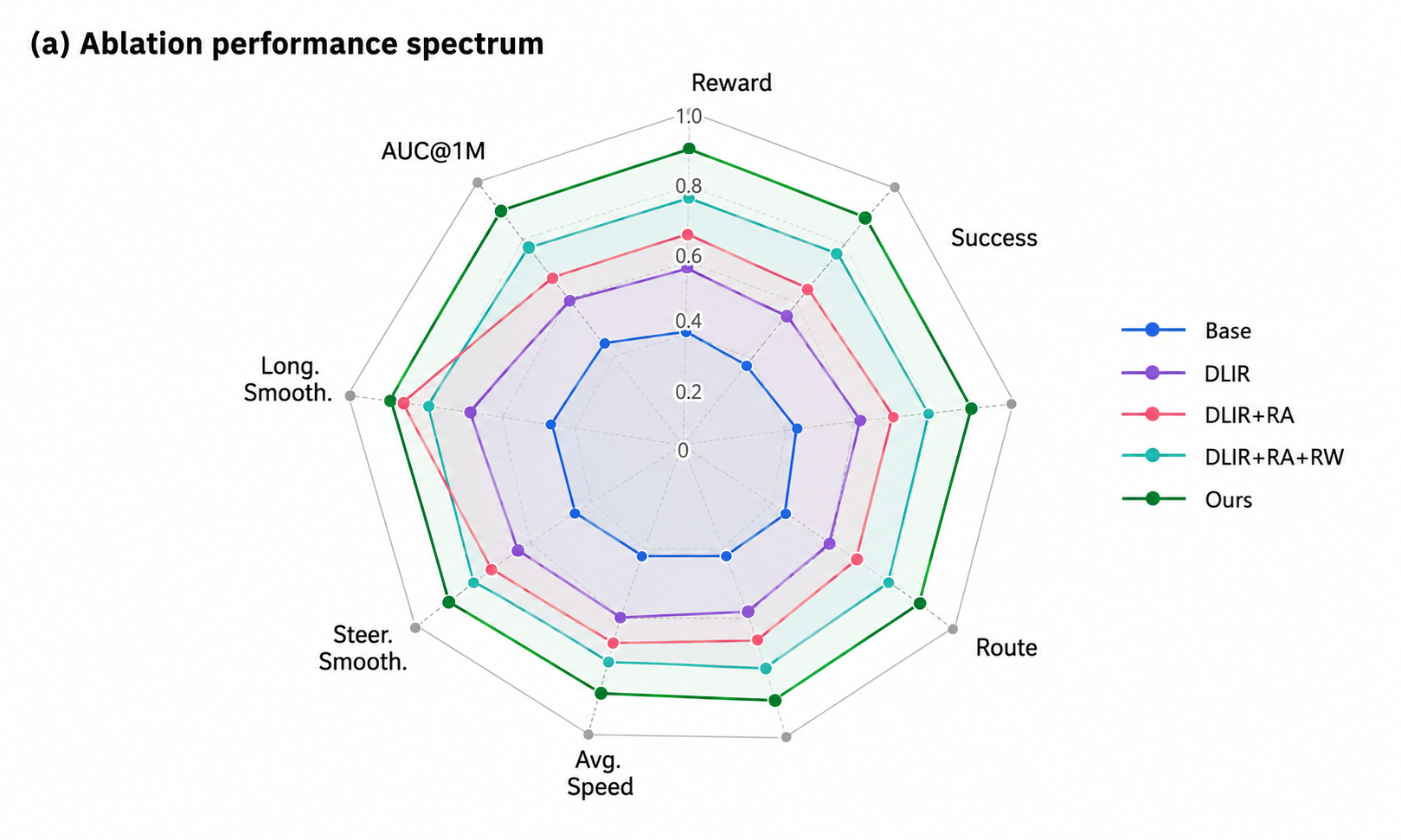}
    \caption{Ablation spectrum of the proposed framework. The full model maintains a stronger balance across task, safety, and smoothness metrics than variants with removed risk-aware generation or reward weighting components.}
    \label{fig:ablation}
\end{figure}

The ablation spectrum in Fig.~\ref{fig:ablation} compares the full method with variants that remove or replace key components. The full model provides a stronger overall balance across reward, success, route completion, average speed, steering smoothness, longitudinal smoothness, and area-under-curve performance. Variants without risk-aware generation or risk-field weighting lose performance on multiple axes, which supports the claim that the framework benefits from both scenario-level and reward-level risk modelling.

The ablation result also shows why the proposed design should be evaluated as a loop rather than as a single reward term. Scenario generation changes the distribution of training and test cases. Risk-field reward changes how the policy responds inside each case. Closed-loop validation connects both effects by feeding failures back into the scenario library.

\subsection{Claim-Evidence Summary}

Table~\ref{tab:claim_evidence} summarizes the relationship between the main claims and the evidence used in this paper. This table is included to make the research logic explicit. The framework claim is supported by the closed-loop architecture and layered model. The risk representation claim is supported by the risk field definition and visualization. The training claim is supported by the training curves and ablation spectrum.

\begin{table*}[!t]
\centering
\caption{Claim-evidence alignment for the proposed framework}
\label{tab:claim_evidence}
\footnotesize
\begin{tabular}{@{}L{0.28\textwidth} L{0.24\textwidth} L{0.40\textwidth}@{}}
\toprule
Claim & Evidence & Interpretation \\
\midrule
Closed-loop twin structure supports iterative validation & Figs.~\ref{fig:overall_framework}--\ref{fig:validation_flow} & Physical data, virtual reconstruction, policy testing, and scenario updates are connected. \\
Risk fields provide dense safety guidance & Fig.~\ref{fig:risk_field} and reward design & Spatial risk is available before collision, unlike sparse terminal penalties. \\
Risk-aware scenarios improve validation pressure & Fig.~\ref{fig:scenario_generation} & Scenarios are ranked by risk rather than sampled only by appearance. \\
The full framework improves balanced behaviour & Figs.~\ref{fig:training_curves} and \ref{fig:ablation} & Training and ablation trends support improvements across safety and task metrics. \\
\bottomrule
\end{tabular}
\end{table*}

Overall, the evidence suggests that explicit risk modelling improves the usefulness of digital twin-based validation. The method is not only a visualization framework. It uses risk information to select scenarios, shape policy learning, and organize validation results. This makes the digital twin more actionable for autonomous driving safety analysis.

\section{Conclusion}
\label{sec:conclusion}

This paper presents a risk-field enhanced closed-loop digital twin framework for autonomous driving safety validation. The framework connects physical data acquisition, virtual scenario reconstruction, autonomous driving algorithm evaluation, risk-aware scenario generation, reinforcement learning policy training, and scenario-library evolution. A driving risk field is introduced as a unified representation that links scenario ranking, safety evaluation, and risk-guided learning.

The simulation-style evaluation shows that the proposed framework can organize validation evidence around training behaviour, route completion, collision avoidance, and control smoothness. The training curves indicate improved stability, while the ablation spectrum suggests that combining risk-aware scenario generation with risk-field reward shaping provides more balanced behaviour than using risk-agnostic baselines or isolated risk penalties.

The boundary of this study should also be clear. The credibility of the framework depends on vehicle dynamics fidelity, sensor and traffic-agent modelling, risk-field calibration, and sim-to-real transfer. If the virtual twin underestimates perception noise or interaction uncertainty, the resulting safety conclusion may become overconfident. Future work should therefore focus on uncertainty-aware risk fields, high-fidelity sensor twins, calibrated criticality metrics, standardized scenario descriptions, and validation with real driving logs. A useful next step is to combine closed-loop digital twins with real fleet data so that virtual validation can be repeatedly checked against observed disengagements, near misses, and human takeover events.

\end{document}